\documentclass[]{article}
\usepackage{fullpage}
\usepackage{amsfonts}

\usepackage{graphicx,times,amsmath}

\begin{document}

\title{N2Sky - Neural Networks as Services in the Clouds}

\author{Erich Schikuta and Erwin Mann\\
  University of Vienna, Faculty of Computer Science \\
  Research Group Workflow Systems and Technology \\
  W\"ahringerstr. 29, A-1090 Vienna, Austria\\
  email: erich.schikuta@univie.ac.at}
%

\date{}

\maketitle

\begin{abstract}
We present the N2Sky system, which provides a framework for the exchange
of neural network specific knowledge, as neural network paradigms and objects,
by a virtual organization environment. It follows the sky computing paradigm delivering
ample resources by the usage of federated Clouds.
N2Sky is a novel Cloud-based neural network simulation
environment, which follows a pure service oriented approach.
The system implements a transparent environment aiming to enable both novice
and experienced users to do neural network research easily and comfortably.
N2Sky is built using the RAVO reference architecture of virtual organizations
which allows itself naturally integrating into the Cloud service stack
(SaaS, PaaS, and IaaS) of service oriented architectures.

\end{abstract}
\section{Introduction}
A Virtual Organisation is a logical orchestration of
globally dispersed resources to achieve common goals \cite{FosterFall2001}.
It couples a wide variety of geographically distributed computational resources (such as PCs, workstations and supercomputers), storage systems, databases, libraries and special purpose scientific instruments to present them as a unified integrated resource that can be shared transparently by communities.

We are living in the era of virtual collaborations, where resources are logical and solutions are virtual. Advancements
on conceptual and technological level have enhanced the way people communicate.
The exchange of information and resources between researchers is one driving stimulus for development. This is just as valid for the neural information processing community as for any other research community. As described by the UK e-Science initiative \cite{UKeS} several goals can be reached by the usage of new stimulating techniques, such as enabling more effective and seamless collaboration of dispersed communities, both scientific and commercial, enabling large-scale applications and transparent access to \hbox{''}high-end\hbox{``} resources from the desktop, providing a uniform \hbox{''}look \& feel\hbox{``} to a wide range of resources and location independence of computational resources as well as data.

In the Computational Intelligence community these current developments are not used to the maximum possible extent until now. As an illustration for this we highlight the large number of neural network simulators that have been developed, as for instance the Self-Organizing Map Program Package (SOM-PAK)~\cite{cite:SOM} and the Stuttgart Neural Network Simulator (SNNS)~\cite{cite:SNNS} to name only a few. Many scientists, scared of existing programs failing to provide an easy-to-use, comprehensive interface, develop systems for their specific neural network applications. This is also because most of these systems lack a generalized framework for handling data sets and neural networks homogeneously. This is why we believe that there is a need for a neural network simulation system that can be accessed from everywhere.

We see a solution to this problem in the N2Sky system.
Sky Computing is an emerging computing model where resources from multiple Cloud providers are leveraged to create large scale distributed infrastructures. The term \emph{Sky Computing} was coined in \cite{skycomputing09} and was defined as an architectural concept that denotes federated Cloud computing. It allows for the creation of large infrastructures consisting of Clouds of different affinity, i.e. providing different types of resources, e.g. computational power, disk space, networks, etc., which work together to form one giant Cloud or, so to say, a \emph{sky computer}.

N2Sky is an artificial neural network simulation environment providing functions like creating, training and evaluating neural networks. The system is Cloud based in order to allow for a growing virtual user community. The simulator interacts with Cloud data resources (i.e. databases) to store and retrieve all relevant data about the static and dynamic components of neural network objects and with Cloud computing resources to harness free processing cycles for the \hbox{''}power-hungry\hbox{``} neural network simulations. Furthermore the system allows to be extended by additional neural network paradigms provided by arbitrary users.

The layout of the paper is as follows: In the following section we give
the motivation behind the work done. In section \ref{sec:design} we present the design principles behind the N2Sky development. The system deployment within a Cloud environment is described in section \ref{sec:deployment}. The architecture of N2Sky is laid out in section \ref{sec:architecture}. In section \ref{sec:interface} the interface of N2Sky is presented.
The paper closes with a look at future developments and research directions in Section 5.

\section{Towards a Cloud-Based ANN Simulator}
%

Over the last few years, the authors have developed several neural network simulation
systems, fostered by current computer science paradigms.

NeuroWeb \cite{cite:NeuroWeb} is a simulator for neural networks
which exploits Internet-based networks as a transparent layer to
exchange information (neural network objects, neural network
paradigms). NeuroAccess \cite{cite:NeuroAccess} deals with the
conceptual and physical integration of neural networks into
relational database systems. The N2Cloud system \cite{Huqqani2010} is based on a
service oriented architecture (SOA) and is a further evolution
step of the N2Grid systems \cite{schiki02}. The original idea behind the N2Grid system was
to consider all components of an artificial neural network as data
objects that can be serialized and stored at some data site in the Grid, whereas
N2Cloud will use the storage services provided by the Cloud
environment. This concept covers not only cloud storage but also all heterogeneous date sources available on the web~\cite{schiki03,schiki08}.

The motivation to use Cloud technology lies in the essential characteristics of this model for
enabling ubiquitous, convenient, on-demand network access to a shared pool of configurable computing resources (e.g., networks, servers, storage, applications, and services) that can be rapidly provisioned and released with minimal management effort or service provider interaction \cite{NISTCloudDefinition11}.

In the light of the development of N2Sky and the goal to develop a virtual organisation for the
neural network community five Cloud characteristics \cite{NISTCloudDefinition11} can be revisited by the following:
\begin{itemize}
  \item \textbf{Shared Pool of Resources.} Resources are shared by multiple tenants. A tenant is defined by the type of cloud being used; therefore a tenant can be either a department, organization, institution, etc.

      N2Sky shares besides hardware resources also knowledge resources. This allows the creation of a shared pool of neural net paradigms, neural net objects and other data and information between researchers, developers and end users worldwide.

  \item \textbf{On-demand self-service.} Consumers can create their computing resources (software, operating system, or server) within mere minutes of deciding they need it without requiring human interaction with each service provider.

      N2Sky allows for transparent access to "high-end" resources (computing and knowledge resources) stored within the Cloud on a global scale from desktop or smart phone, i.e. whenever the consumer needs it independently from the consumer local infrastructure situation.

  \item \textbf{Broad network access.} Users can access the computing resources from anywhere they need it as long as they are connected to the network.

      N2Sky fosters location independence of computational, storage and network resources.

  \item \textbf{Rapid elasticity.} Computing resources can scale up or scale down based on the users needs. To end users this appears to be unlimited resources.

      N2Sky delivers to the users a resource infrastructure which scales according to the problem. This leads to the situation that always the necessary resources are available for any neural network problem.

  \item \textbf{Measured service.} Services of cloud systems automatically control and optimize resource use by leveraging a metering capability enabling the pay-as-you-go model. This allows consumers of the computing resources to pay based on their use of the resource.

      N2Sky supports the creation of neural network business models. Access to neural network resources, as novel paradigms or trained neural networks for specific problem solutions, can be free or following certain business regulations, e.g. a certain fee for usage or access only for specific user groups.
\end{itemize}

The presented N2Sky environment takes up the technology of N2Cloud to a
new dimension using the virtual organisation paradigm. Hereby the RAVO
reference architecture is used to allow the easy integration of N2Sky into the
Cloud service stack using SaaS, PaaS, and IaaS.
Cloud computing is a large scale distributed computing paradigm
for utility computing based on virtualized, dynamically scalable
pool of resources and services that can be delivered on-demand
over the Internet. In the scientific community
it is sometimes stated as the natural evolution of Grid computing. Cloud computing
therefore became a buzz word after IBM and Google collaborated in
this field followed by IBM's ''Blue Cloud"
\cite{cite:IBMBlueCloud} launch. Three categories can be
identified in the field of Cloud computing:

\begin{itemize}
  \item \textbf{Software as a Service (SaaS).} This type of Cloud
  delivers configurable software applications offered by third
  party providers on an on-demand base and made  available to
  geographically distributed users via the Internet. Examples are
  Salesforce.com, CRM, Google Docs, and so on.
  \item \textbf{Platform as a Service (PaaS).} Acts as a
  runtime-system and application framework that presents itself as
  an execution environment and computing platform. It is
  accessible over the Internet with the sole purpose of acting as
  a host for application software. This paradigm offers customers
  to develop new applications by using the available development
  tools and API's. Examples are Google's App engine and
  Microsoft's Azure, and so on.
  \item \textbf{Infrastructure as a Service (IaaS).} Traditional
  computing resources such as servers, storage, and other forms of
  low level network and physical hardware resources are hereby
  offered in a virtual, on-demand fashion over the Internet. It
  provides the ability to provide on-demand resources in specific
  configurations. Examples include Amazon's EC2 and S3, and so on.
\end{itemize}


\section{N2Sky Design}\label{sec:design}

Information Technology (IT) has become an essential part of our daily life. Utilization
of electronic platforms to solve logical and physical problems is extensive. Grid
computing \cite{FosterFall2001} is often related with Virtual Organisations (VOs) when it comes to creation of an E-collaboration.
The layered architecture for Grid computing has remained ideal for VOs.

However, the Grid computing paradigm has some limitations. Existing Grid environments
are categorized as data grid or computational grid. Today, problems being
solved using VOs require both data and storage resources simultaneously. Scalability
and dynamic nature of the problem solving environment is another serious
concern. Grid computing environments are not very
flexible to allow the participant
entities enter and leave the trust. Cloud computing seems to be a promising
solution to these issues. Only, demand driven, scalable and dynamic problem solving
environments are target of this newborn approach. Cloud computing is not
a deviation concept from the existing technological paradigms, rather it is an evolution.
Cloud computing centers around the concept of "Everything as a Service" (XaaS), ranging from
hardware/software, infrastructure, platform, applications and even humans are configured
as a service. Most popular service types are IaaS, PaaS and SaaS.

Existing paradigms and technology are used to form VOs, but lack
of standards remained a critical issue for the last two decades.
Our research endeavor
focused on developing  a Reference Architecture for Virtual Organizations (RAVO)  \cite{disskhalil}.
It is intended as a standard for building Virtual Organizations (VO). It gives
a starting point for the developers, organizations and individuals to collaborate electronically for achieving
common goals in one or more domains. RAVO consists of two parts,
\begin{enumerate}
  \item The requirement analysis phase, where boundaries of the VO are defined and components are identified.
A gap analysis is also performed in case of evolution (up-gradation) of an existing system to a VO.
  \item The blueprint for a layered architecture, which defines mandatory and optional components of the VO.
\end{enumerate}
This approach allows to foster new
technologies (specifically the SOA/SOI paradigm realized by Clouds) and the extensibility and changeability of the VO to be developed.

\begin{figure}[htp]
  \begin{center}
    \includegraphics[width=1.0\columnwidth]{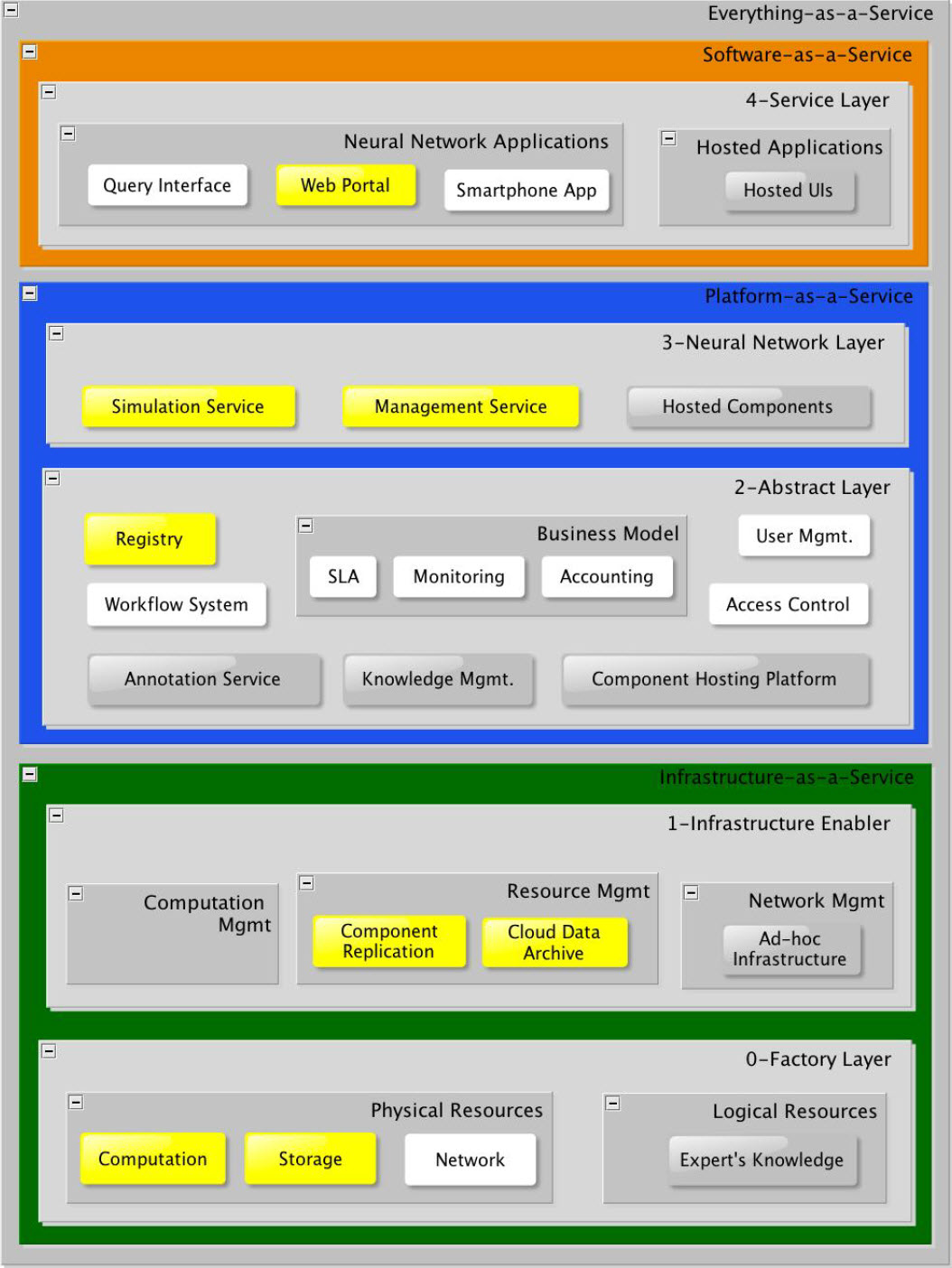}
  \end{center}
  \caption{N2Sky design based on RAVO}
  \label{fig:RavoN2Sky}
\end{figure}

The basic categorization of the the N2Sky design depends on the three layers of the Cloud service stack as they are: Infrastructure as a Service (IaaS), Platform as a Service (PaaS) and Software as a Service (SaaS).
Figure \ref{fig:RavoN2Sky} depicts the components of the N2Sky framework, where yellow components are mandatory, and white and grey components are optional.

\textbf{Infrastructure as a Service (IaaS)} basically provides enhanced virtualisation capabilities. Accordingly, different resources may be provided
via a service interface. In N2Sky the IaaS layer consists of two sub-layers: a Factory layer and an
Infrastructure Enabler Layer. Users need administrative rights for accessing the resources in layer 0 over the
resource management services in layer 1.
\begin{itemize}
  \item Factory Layer (Layer 0). contains physical and logical resources for the N2Sky. Physical resources
comprise of hardware devices for storage, computation cycles and network traffic in a distributed manner.
Logical resources contain expert’s knowledge helping solving special problems like the Paradigm Matching.
  \item Infrastructure Enabler Layer (Layer 1). allows access to the resources provided by the Factory layer. It
consists of protocols, procedures and methods to manage the desired resources.
\end{itemize}

\textbf{Platform as a Service (PaaS)} provides “computational resources via a platform upon which applications and services can be developed
and hosted. PaaS typically makes use of dedicated APIs to control the behaviour of a server hosting engine
which executes and replicates the execution according to user requests. It provides transparent access
to the resources offered by the IaaS layer and applications offered by the SaaS layer. In N2Sky it is divided into two sublayers:
\begin{itemize}
  \item Abstract Layer (Layer 2). This layer contains domain-independent tools that are designed not only for
use in connection with neural networks.
  \item Neural Network Layer (Layer 3). This layer is composed of domain-specific (i.e. neural network) applications.
\end{itemize}

\textbf{Software as a Service (SaaS)} offers “implementations of specific business functions and business processes that are provided with
specific Cloud capabilities, i.e. they provide applications / services using a Cloud infrastructure or platform,
rather than providing Cloud features themselves”. In context of N2Sky, SaaS is composed of one layer, namely the Service Layer.
\begin{itemize}
  \item Service Layer (Layer 4). This layer contains the user interfaces of applications provided in Layer 3 and
is an entry point for both end users and contributors. Components are hosted in the Cloud or can be
downloaded to local workstations or mobile devices.
\end{itemize}

Each of the five layers provide its functionality in a pure service-oriented manner so we can say that N2Sky
realizes the Everything-as-a-Service (XaaS) paradigm.

\section{N2Sky Cloud Deployment}\label{sec:deployment}

At the moment N2Sky facilitates Eucalyptus \cite{cite:eucalyptus}, which is an open source software platform
that implements a Cloud infrastructure (similar to Amazon's
Elastic Compute Cloud) used within a data center. Eucalyptus
provides a highly robust and scalable Infrastructure as a Service
(IaaS) solution for Service Providers and Enterprises.
A Eucalyptus Cloud setup consists of three components the Cloud
controller (CLC), the cluster controller(s) (CC) and node
controller(s) (NC). The Cloud controller is a Java program that,
in addition to high-level resource scheduling and system
accounting, offers a Web services interface and a Web interface to
the outside world. Cluster controller and node controller are
written in the programming language C and deployed as Web services
inside an Apache environment.

Communication among these three types of components is
accomplished via SOAP with WS-Security. The N2Sky System itself is
a Java-based environment for the simulation and evaluation of
neural networks in a distributed environment. The Apache Axis
library and an Apache Tomcat Web container are used as a hosting
environment for the Web Services. To access these services Java
Servlets/JSPs have been employed as the Web frontend.

This design approach allows easy portability of N2Sky to other Cloud platforms.
We just finished a deployment of N2Sky onto the OpenStack environment \cite{cite:openstack}.
The motivation for this move is the change in the policy of Eucalyptus towards a stronger commercial
orientation and the increase in popularity of OpenStack within the Cloud community.
We are also working on a port to OpenNebula \cite{cite:opennebula}, the flagship Cloud project of the European union.
All these ports are simple for the N2Sky software. The effort of porting N2Sky lies in the
fact getting competence into the new Cloud environment.

\begin{figure*}[htbp]
  \begin{center}
    \includegraphics[width=1.0\textwidth]{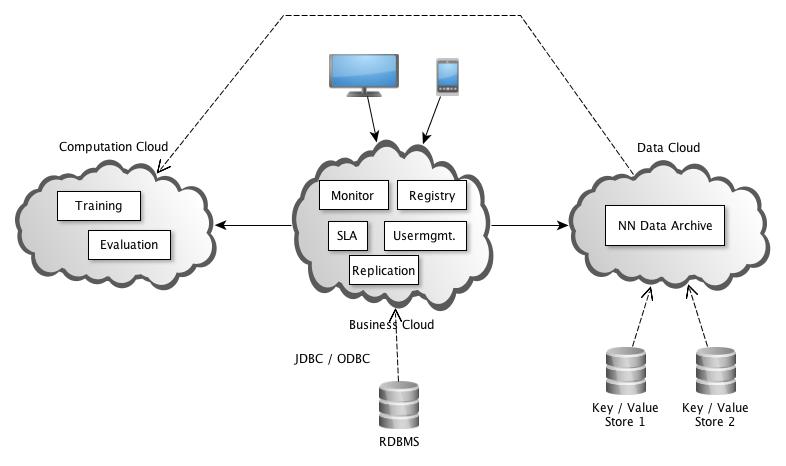}
  \end{center}
  \caption{N2Sky Cloud Deployment}
  \label{deployfig}
\end{figure*}

We maintain that the N2Sky system can be deployed on various Cloud platforms of the underlying infrastructure.
This allows naturally to implement a federated Cloud model, by fostering the specific affinities (capabilities) \cite{jha2009using} of
different Cloud providers (e.g. data Clouds, compute Clouds, etc.). A possible specific deployment is show in
Figure~\ref{deployfig}. Three different Clouds are depicted providing unique capabilities: The Cloud on the left hand side is a computing Cloud, providing strong computing capabilities, responsible for the time consuming training an devaluation phases of neural networks. The Cloud on the right hand side is a data Cloud, which offers extensive storage resources, e.g. by access to relational or NoSQL database systems. The center Cloud is the administrative Cloud, which does not provide specific hardware resources but acts as central access point for the user and acts as mediator to the N2Sky environment, e.g. by applying business models.

\section{N2Sky Architecture}\label{sec:architecture}

\begin{figure*}[htbp]
  \begin{center}
    \includegraphics[width=1.0\textwidth]{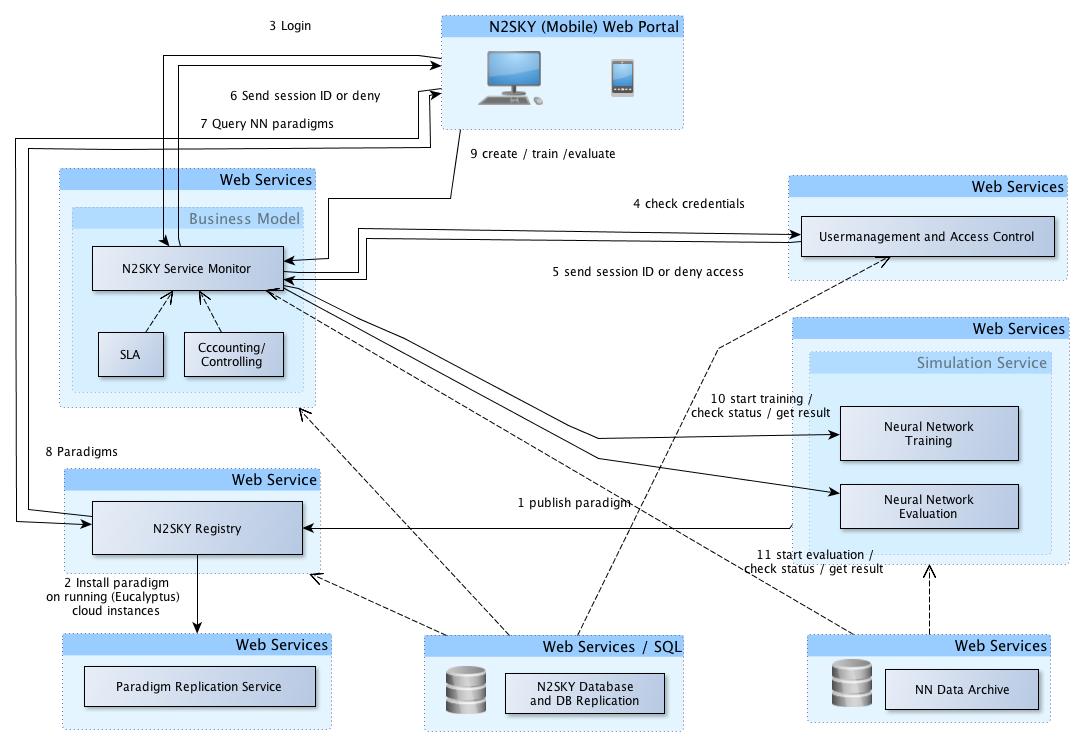}
  \end{center}
  \caption{N2Sky Architecture and Components}
  \label{n2skyfig}
\end{figure*}

The whole system architecture and its components are depicted in
Figure~\ref{n2skyfig}.

A neural network has to be configured or trained (supervised or
unsupervised) so that it may be able to adjust its weights in such
a way that the application of a set of inputs produces the desired
set of outputs. By using a particular paradigm selected by the
user the \textbf{N2Sky Simulation Service} allows basically three
tasks: \textbf{train} (the training of an untrained neural network),
\textbf{retrain} ´(training of a previously trained network again in
order to increase the training accuracy), \textbf{evaluate}
(evaluating an already trained network).
The \textbf{N2Sky Data Archive} is responsible to provide access to data of different
objects (respectively paradigms) of neural networks by archiving
or retrieving them from a database storage service. It can also
publish evaluation data. It provides the method \textbf{put} (inserts data into a data source) and
\textbf{get} (retrieves data from a data source).
The main objective of the \textbf{N2Sky Database Service} is to facilitate users to benefit
from already trained neural networks to solve their problems. So
this service archives all the available neural network objects,
their instances, or input/output data related to a particular
neural network paradigm. This service dynamically updates the
database as the user gives new input/output patterns, defines a
new paradigm or evaluates the neural network.
The \textbf{N2Sky Service Monitor} keeps tracks of the available
services, publishes these services to
the whole system. Initially users interact with it by selecting
already published paradigms like Backpropagation, Quickpropagation, Jordan etc. or submit jobs by defining own
parameters. This module takes advantage of virtualization and
provides a transparent way for the user to interact with the
simulation services of the system.
It also allows to implement business models by an accounting functionality
and restricting
access to specific paradigms.
The \textbf{N2Sky Paradigm/Replication Service} contains the paradigm implementation that can be seen as the
business logic of a neural network service implementation.
The \textbf{N2Sky Registry} administrates the stored neural network paradigms.
The main purpose of N2Sky system is to provide neural network data
and objects to users. Thus the \textbf{N2Sky Java Application/Applet}
provides a graphical user interface (GUI) to the user. It
especially supports experienced users to easily run their
simulations by accessing data related neural network objects that
has been published by the N2Sky service manager and the N2Sky
data service. Moreover the applet provides a facility to end-users
to solve their problems by using predefined objects and paradigms.
For the purpose of thin clients a simple Web browser, which can
execute on a PC or a smart phone, can be used to access the
front-end, the \textbf{N2Sky (Mobile) Web Portal}. It is relying on the \textbf{N2Sky User management Service}
which grants access to the system.

Based on this service layout the following exemplary execution workflow can be derived (the numbers refer to the labels in Figure~\ref{n2skyfig}):

\begin{enumerate}

\item The developer publishes a paradigm service to N2Sky.

\item During paradigm service activation the paradigm is replicated to all running instances (e.g. a Java
web archive is deployed to all running application server instances).

\item Users log in per (mobile) web browser per RESTful web service.

\item Central monitor service dispatches login requests to user management and access control component per
RESTful web service.

\item Callback to service monitor either sending a new session id or deny access.

\item Callback to (mobile) web browser redirecting session id or deny access.

\item Query registry for neural network paradigms per RESTful web service.

\item Callback to (mobile) web browser by sending paradigm metadata.

\item Create a new neural object by using selected paradigm, start new Cloud node instance, start
training and after them start a new evaluation by using training result.

\item Start a new training thread, then get
result and store it over data archive in database.

\item Start a new evaluation thread, then
return result and store it in data archive to database. It is a design goal that no simulation service needs
database connection. These services are able to run on an arbitrary node without having to
deal with database replication.

\end{enumerate}

\section{N2Sky Interface}\label{sec:interface}

The design of the user interface of N2Sky is driven by the following guiding principles:

\begin{itemize}

\item \textbf{Acceptance.} To be accepted by the user the system has to
provide an intuitive and flexible interface with all
necessary (computing and knowledge) resources easily at hand.

\item \textbf{Simplicity.} The handling of a neural network has
to be simple. The environment has to supply functions
to manipulate neural networks in an easy and
(more important) natural way. This can only be
achieved by a natural and adequate representation
of the network in the provided environment of the
system but also by an embedding of the network into
the conventional framework of the system. A neural
network software instantiation has to look, to react
and to be administrated as any other object. The
creation, update and deletion of a neural network
instantiation has to be as simple as that of a conventional
data object.

\item \textbf{Originality.} A neural network object has to be simple,
but it has not to loose its originality. A common
framework always runs the risk to destroy the original
properties of the unique objects. This leads to
the situation that either objects of different types
loose their distinguishable properties or loose their
functionality. A suitable framework for neural networks
has to pay attention to the properties and to
the functional power of neural networks and should
give the user the characteristics he is expecting.

\item \textbf{Homogeneity.} Neural networks have to be considered
as conventional data, which can be stored in any data
management environment, as database system or the
distributed data stores in the Cloud.
From the logical point of view a neural network
is a complex data value and can be stored as a normal
data object.

\item \textbf{System extensibility.} N2Sky offers an easy
to use interface for neural network researchers to extend
the set of neural network paradigms. This can be done by
accessing paradigms from a N2Sky paradigm service,
or by uploading new paradigms. A new
paradigm can be both easily integrated into the system
and provided to other researcher on the Grid by storing
it on the paradigm server.

\end{itemize}

\begin{figure}[htbp]
  \centering
      \includegraphics[width=1.0\columnwidth]{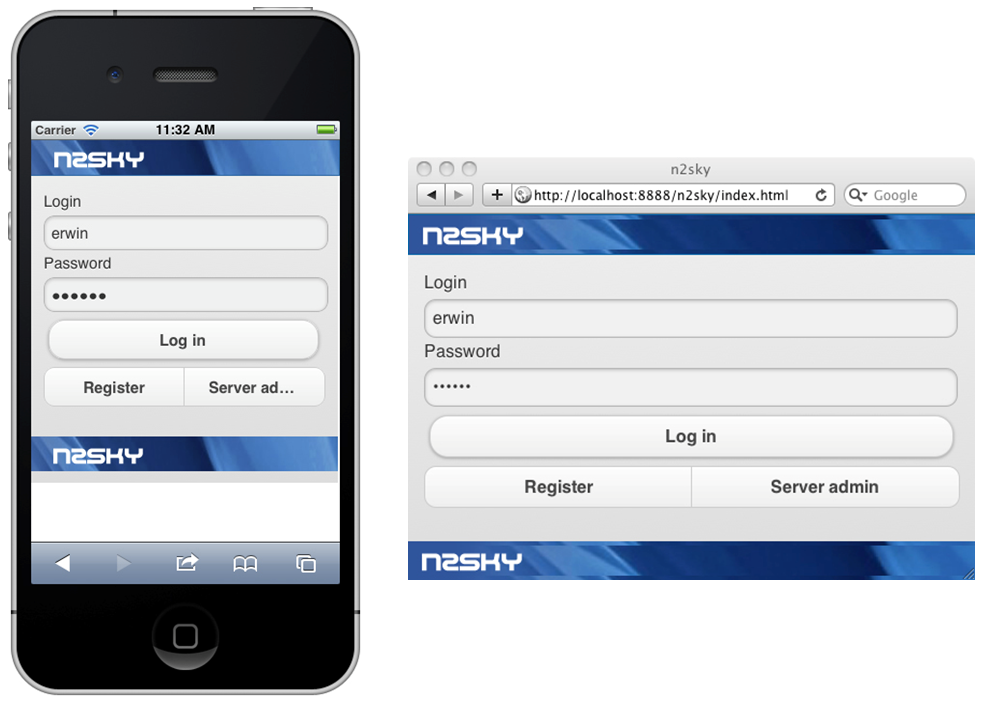}
    \caption{N2Sky Login Screens, iPhone (left) and Mac (right)\label{login}}
\end{figure}

Following these principles we developed a portable interface, which self-adapts to different user platforms.
Technically speaking, we based the development of the N2Sky interface on HTML5, CSS 3.0 and JQuery. Thus
the user needs a conventional web browser only (as Safari, Chrome, Mozilla Firefox, or Internet Explorer) to access the N2Sky portal and can use arbitrary user platforms, as workstations, PCs, Macs, tablets, or smart phones.
In Figure \ref{login} the N2Sky login screen is shown on an iPhone and a Mac as example for this portability issue.

\subsection{Interface Walk-Through}

Basically N2Sky provides screens for the classical neural network tasks. In the following we present a short walk-through of the training of a Backpropagation network. In this presentation we show the screenshots of an iPhone only.

\begin{itemize}
  \item \textbf{Paradigm subscription.} The user chooses an published available neural network paradigm on the N2Sky paradigm server and instantiates a new neural network object based on this paradigm. In Figure \ref{subscription} the most important steps of this workflow are shown (from left to right).

      First the paradigm is chosen from paradigm store on the paradigm service by using a SQL query statement. Then the information on the chosen paradigm is shown to the user by the Vienna Neural Netwrok specification language \cite{cite:ViNNSL08}. After deciding for an appropriate paradigm the user can instantiate a new network object. In the shown example the user creates a three layer, fully connected Backpropagation network.

\begin{figure}[htbp]
  \centering
      \includegraphics[width=1.0\columnwidth]{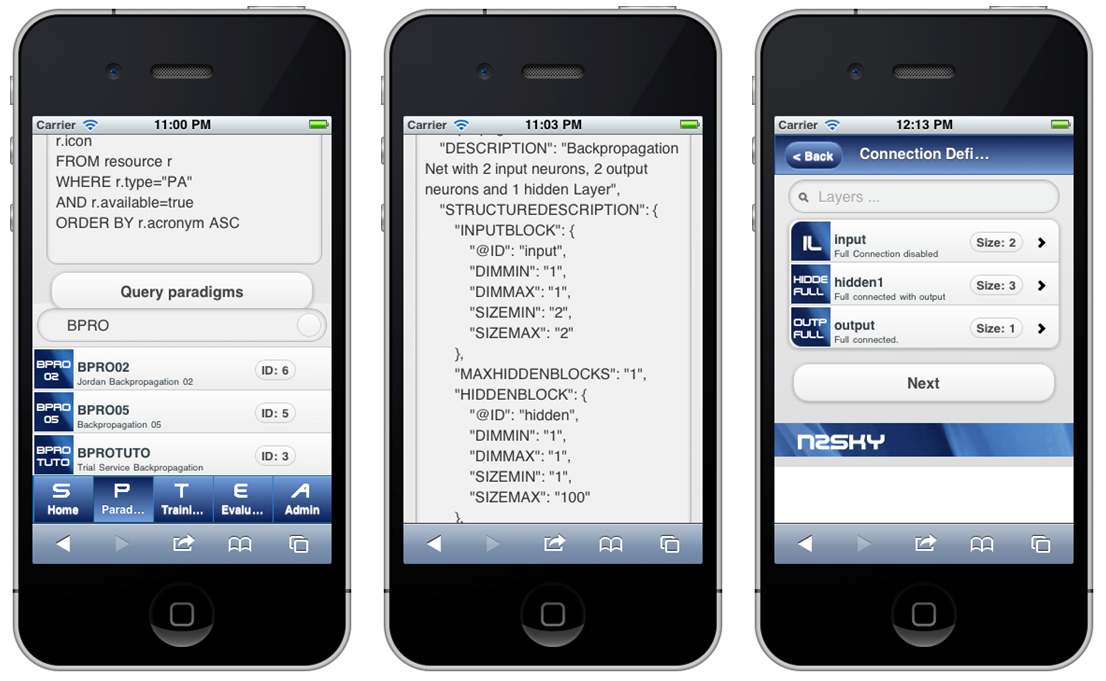}
    \caption{Paradigm subscription\label{subscription}}
\end{figure}

  \item \textbf{Training.} The user starts a training phase on the newly created Backpropagation network. First the user specifies the training parameters (as momentum term, activation function, etc.). Then the training data set (training patterns) have to defined. Hereby the N2Sky allows both the explicit specification of the data used (as shown in the Figure) and the specification of the dataset by a SQL or NoSQL query statement (see next section for more details). The training is started in the Cloud and the error graph is shown on the iPhone. After completion of the training the training results, (e.g. calculated weight values, error values, etc.), are shown and stored in the database for further usage (e.g. evaluation).

\begin{figure}[htbp]
  \centering
      \includegraphics[width=1.0\columnwidth]{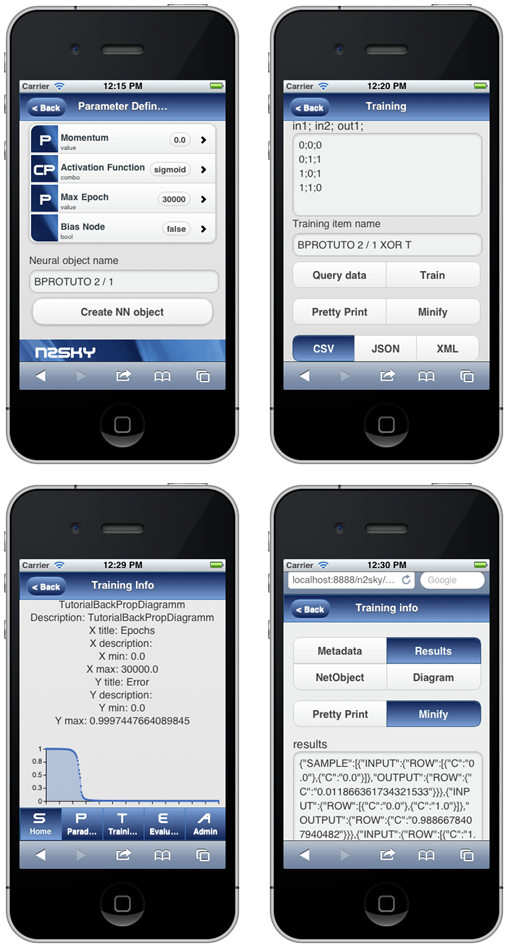}
    \caption{Network training\label{training}}
\end{figure}

  \item \textbf{Evaluation.} The last step is the evaluation of the trained neural networks for problem solution. An evaluation object is created, which is using an existing training object. Also here the user has the possibility to define the evaluation data by an explicit list or a query statement. After the evaluation is performed in the Cloud the finalized evaluation object (the solution to the given problem) can be used elsewhere. Due to the situation that it is stored in the data store, it is available and accessible from everywhere.

\begin{figure}[htbp]
  \centering
      \includegraphics[width=1.0\columnwidth]{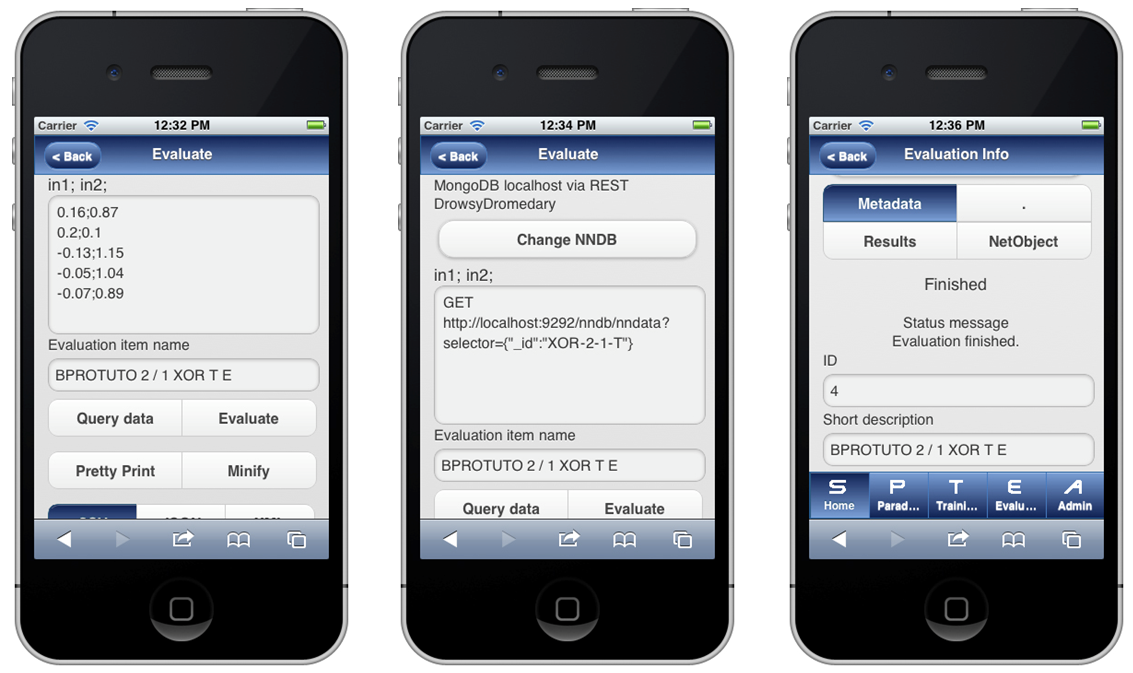}
    \caption{Network evaluation\label{evaluation}}
\end{figure}

\end{itemize}

\subsection{Datastream Concept}

A highlight of the N2Sky system is the use of standardized and user-friendly
database query languages for searching of network paradigms and objects and defining the training
and evaluation data set.
The functional data stream definition allows to specify
the data sets in a comfortable and natural way. It is not necessary
to specify the data values explicitly, but the
data streams can be described by SQL and NoSQL statements.So it is easily
possible to use ’real world’ data sets as training
or evaluation data on a global scale.
This approach implements interface homogeneity to the user too, who applies the the same tool both for administration tasks (choosing a neural network paradigm or object) and the analysis of the stored information.

Specifically for the definition of the training and evaluation data set, which can be huge data volumes,
this functional specification by a query language statement is extremely comfortable. This unique feature allows for combining globally stored, distributed data within the N2Sky environment easily.

\begin{figure}[htbp]
  \centering
      \includegraphics[width=1.0\columnwidth]{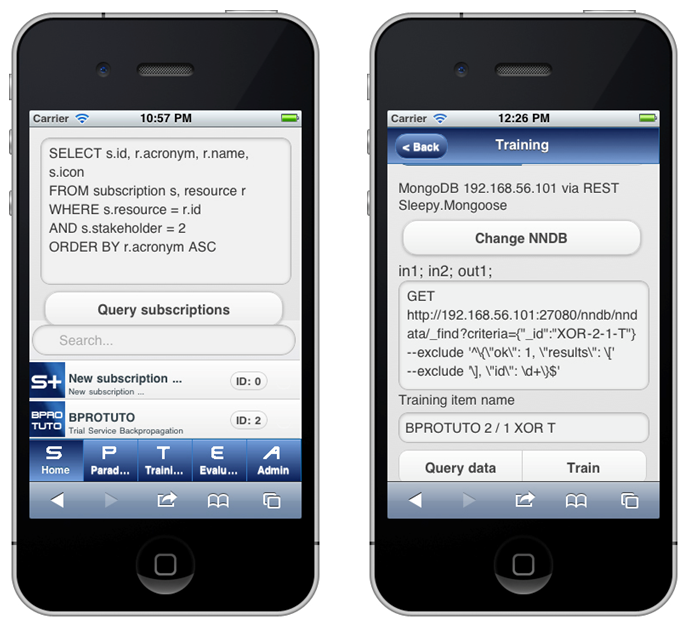}
    \caption{SQL and NoSQL datastream definition\label{datastream}}
\end{figure}

In Figure \ref{datastream} examples for SQL using a relational database system (left) and NoSQL using MongoDB (right) are shown. This Figure also exemplifies the usage of this approach on the one hand for administrative tasks (choosing a neural network paradigm on the paradigm server) and on the other hand for data set specification (accessing big data on an Internet NoSQL data store via a RESTful access mechanism)

\section{Summary}

In this paper we presented N2Sky, a Cloud-based
framework enabling the Computational Intelligence
community to share and exchange the neural network resources
within a Virtual Organisation.
N2Sky is a prototype system with quite some room for further
enhancement. Ongoing research is done in the following areas:

\begin{itemize}
  \item We are working on an enhancement of the neural network
  paradigm description language ViNNSL \cite{cite:ViNNSL08} to
  allow for easier sharing of resources between the paradigm
  provider and the customers. We are also aiming to build a
  generalized semantic description of resources for exchanging
  data.
  \item Parallelization of neural network training is a further
  key for increasing the overall performance. Based on our
  research on neural network parallelization \cite{cite:NNpar04,schiki01,schiki11}
  we envision an automatically definition and usage of
  parallelization patterns for specific paradigms. Furthermore the
  automatic selection of capable resources in the Cloud for
  execution, e.g. multi-core or cluster systems is also a hot
  topic within this area.
  \item Key for fostering of cloud resources are service level agreements (SLAs) which
  give guarantees on quality of the delivered services. We are working on the 
  embedment of our research findings on SLAs~\cite{schiki04,schiki10,schiki12} into N2Sky to allow for novel business models~\cite{schiki05,schiki06,schiki07,schiki09} on the selection and usage of neural network resources based on quality of service attributes~\cite{schiki14}.
  \item A further important issue is to find neural network solvers for
  given problems, similar to a "Neural Network Google". In the
  course of this research we are using ontology alignment by
  mapping problem ontology onto solution ontology.
\end{itemize}



\end{document}